# STDHL: Spatio-Temporal Dynamic Hypergraph Learning for Wind Power Forecasting

Xiaochong Dong, *Member, IEEE*, Xuemin Zhang, *Member, IEEE*, Ming Yang, *Senior Member, IEEE*, Shengwei Mei, *Fellow, IEEE*

*Abstract*— Leveraging spatio-temporal correlations among wind farms can significantly enhance the accuracy of ultra-short-term wind power forecasting. However, the complex and dynamic nature of these correlations presents significant modeling challenges. To address this, we propose a spatio-temporal dynamic hypergraph learning (STDHL) model. This model uses a hypergraph structure to represent spatial features among wind farms. Unlike traditional graph structures, which only capture pair-wise node features, hypergraphs create hyperedges connecting multiple nodes, enabling the representation and transmission of higher-order spatial features. The STDHL model incorporates a novel dynamic hypergraph convolutional layer to model dynamic spatial correlations and a grouped temporal convolutional layer for channel-independent temporal modeling. The model uses spatio-temporal encoders to extract features from multi-source covariates, which are mapped to quantile results through a forecast decoder. Experimental results using the GEFCom dataset show that the STDHL model outperforms existing state-of-the-art methods. Furthermore, an in-depth analysis highlights the critical role of spatio-temporal covariates in improving ultra-short-term forecasting accuracy.

## I. INTRODUCTION

Unlike conventional adjustable power sources, wind turbine power generation is highly dependent on meteorological conditions and cannot be controlled at will. Given the inherent uncertainty in weather patterns, accurate wind power forecasting is essential for unit commitment, demand response, and frequency regulation [1-3]. Wind power forecasting is typically categorized three temporal scales: medium-term, short-term, and ultra-short-term [4]. Among these, ultra-short-term forecasting, with the shortest forecast horizon, plays a pivotal role in maintaining real-time supply-demand balance and enabling the seamless integration of wind power into the energy system. The regulations governing ultra-short-term forecasting vary among operators. In China, forecasts are updated every 15 minutes and covers next 4 hours. In Electric Reliability Council of Texas (ERCOT), forecasts are updated every 15 minutes, but covers the next 8 hours [5].

X. Dong, X. Zhang (corresponding author) and S. Mei are with the State Key Laboratory of Power System Operation and Control, Department of Electrical Engineering, Tsinghua University, Beijing, 100084, China (e-mail: dream_dxc@163.com; zhangxuemin@mail.tsinghua.edu.cn; meishengwei@mail.tsinghua.edu.cn).

Ming Yang is with the Key Laboratory of Power System Intelligent Dispatch and Control, Shandong University, Jinan 250061, China (e-mail: myang@sdu.edu.cn).

From the perspective of earth system modeling, the future wind speed at a target location is influenced by the state of atmospheric motion within a spatial range [6]. The information required about the atmospheric state is commonly referred to as the initial field. Solving the set of partial differential equations that govern atmospheric motion allows for the estimation of future wind speeds [7]. However, this approach heavily relies on the completeness of initial field as covariates. Similarly, comprehensive and accurate covariates are essential as inputs for wind power forecasting problem.

In the wind power forecasting, covariates can be categorized into two types: measured data and numerical weather prediction (NWP) data. Measured data include power and meteorological data [8]. Given the spatio-temporal continuity of atmospheric flow, power and meteorological data collected from different wind farms within a region can serve as valuable references for ultra-short-term power forecasting. However, measured data are limited in terms of sampling points and geographical coverage, making extrapolations prone to lag and error accumulation [9].

NWP data are generated through numerical calculations based on the global initial field [10]. When the initial field are complete, NWP data can effectively infer wind speed trends over the next few days. However, the high computational complexity and significant processing time associated with NWP data reduce timeliness, thereby limiting their suitability for ultra-short-term rolling updates [11].

To improve forecasting performance, it is crucial to combine the strengths of both measured data and NWP data for a complementary effect. Additionally, incorporating data from multiple stations can enhance the comprehensiveness of covariates [12-13].

While multi-source data form the foundation of wind power forecasting, forecast models are the essential tools for achieving accurate forecasts. Forecast models are categorized into physical, statistical, machine learning, and deep learning models. Physical models establish the relationship between wind speed and wind power based on wind turbine power curve but often overlook other influencing factors [14]. Statistical models explicitly model the relationships between various factors and wind power, but struggle to capture nonlinear correlations between meteorological features [15]. Machine learning models often rely on feature engineering to enhance regression performance. However, manually selecting or designing effective features requires extensive prior knowledge, making it challenging to uncover potentially valuable features [16]. In contrast, deep learning models, which use neural networks to represent implicit relationships between meteorological features and wind power, have garnered significant attention for their strong nonlinear mapping capabilities [17].

In terms of multi-source data fusion and feature extraction, spatial feature extraction using neural networks has been a focal point of research. Early studies used convolutional neural networks for spatial domain feature extraction, but these faced limitations in handling non-Euclidean geographical data structures [18-19]. More recently, graph neural networks (GNN) have emerged as a promising solution [20]. GNN describe the spatial features between multiple wind farms using nodes, edges, and an adjacency matrix, where the elements of the adjacency matrix represent the strength of the spatial correlation. Since geographically closer wind farms tend to exhibit higher spatial correlations, functions inversely related to distance can be used to construct adjacency matrix [21]. In addition, Pearson correlation coefficients can also be used to construct adjacency matrices [22]. However, these methods produce symmetric adjacency matrices that can only describe undirected spatial correlations. To address this limitation, some studies have introduced the concept of directed graphs, which can represent directed spatial features between nodes [23-24]. Furthermore, considering the dynamic spatial correlations, the mapping relationship between covariates and the adjacency matrix can be constructed using kernel functions or neural networks [25-26]. However, most existing studies primarily utilize pair-wise graph structures, which are better suited for traffic or power networks with fixed node connections [27-28]. Due to the spatial continuity of atmospheric systems, spatial features are shared among multiple nodes, indicating that higher-order feature transmission occurs across wind farms. To represent higher-order correlations among multiple nodes, a pair-wise graph structure requires more edges to establish node connections or relies on repeated neighborhoods information transfers, which adds additional computational burden [29].

In conclusion, integrating spatio-temporal covariates for ultra-short-term wind power forecasting remains a significant challenge. In this work, we propose the spatio-temporal dynamic hypergraph learning (STDHL) model, which makes the following major contributions:

(1) A novel hypergraph dynamic convolutional mechanism is proposed to model the spatial features among multiple wind farms. This approach enables dynamic modeling of spatial features and facilitates the transfer of shared features across multiple nodes, thereby enhancing the representation of higher-order spatial features.

(2) The STDHL integrates multi-source spatio-temporal covariates through a pair of spatio-temporal encoders and a forecast decoder. By combining stacked dynamic hypergraph convolutional layers with grouped temporal convolutional layers, the model can capture spatial features across multiple time scales while maintaining the ability to independently model temporal features of each nodes.

The rest of this manuscript is organized as follows. Section II presents the formulation of spatio-temporal forecasting. Section III presents the network structure of the STDHL. Section IV summarizes the case study. Section V presents the numerical results. Finally, Section VI concludes the manuscript.

## II. PROBLEM FORMULATION

### A. Spatio-Temporal Forecasting

In the spatio-temporal forecasting problem, the objective is to forecast the power $\widehat{\boldsymbol{y}}_{t_0+1:t_0+T} \in \mathbb{R}^{N \times T}$ of $N$ wind farms over a future time horizon $\mathcal{T} = \{t_0+1,...,t_0+T\}$, where $t_0$ is the current moment. Generally, the covariates for the spatio-temporal forecasting task include measured data and NWP data. Measured data consists of measured power $\boldsymbol{y}_{t_0-T'+1:t_0} \in \mathbb{R}^{N \times T'}$ and measured meteorological data $\boldsymbol{x}_{t_0-T'+1:t_0} \in \mathbb{R}^{F' \times N \times T'}$, where $T'$ is a fixed look back time horizon, and $F'$ is the number of measured meteorological features. NWP data provides forecast meteorological data $\widehat{\boldsymbol{x}}_{t_0-\tau+1:t_0+T+\tau} \in \mathbb{R}^{F \times N \times (T+2\tau)}$, where $F$ is the number of forecast meteorological features, and $\tau$ is the extended time horizon. Here, no distinction is made between the extended time horizon $\tau$ before and after the target time horizon $[t_0+1:t_0+T]$.

For deterministic forecasting, the goal is to find a mapping $f(\cdot)$ that minimizes the mean absolute error (MAE) or root mean squared error (RMSE), bringing the forecast results $\widehat{\boldsymbol{y}}_{t_0+1:t_0+T}$ close to actual values $\boldsymbol{y}_{t_0+1:t_0+T}$.

$$\widehat{\boldsymbol{y}}_{t_0+1:t_0+T} = f(\boldsymbol{y}_{t_0-T'+1:t_0}, \boldsymbol{x}_{t_0-T'+1:t_0}, \widehat{\boldsymbol{x}}_{t_0-\tau+1:t_0+T+\tau}) \quad (1)$$

For quantile probabilistic forecasting, forecast results is the set of quantiles of interest be $\{u_q \mid q=1,...,Q\}$, where $Q$ is the number of quantiles. The task is to find a mapping $g(\cdot)$ that minimizes the quantile loss, bringing the forecast quantiles $\widehat{\boldsymbol{y}}^{u_q}_{t_0+1:t_0+T}$ close to actual values $\boldsymbol{y}_{t_0+1:t_0+T}$.

$$\widehat{\boldsymbol{y}}^{u_q}_{t_0+1:t_0+T} = g(\boldsymbol{y}_{t_0-T'+1:t_0}, \boldsymbol{x}_{t_0-T'+1:t_0}, \widehat{\boldsymbol{x}}_{t_0-\tau+1:t_0+T+\tau}) \quad (2)$$

### B. Graph Structure

The traditional graph structure can be defined as $G=(V,E,A)$, which includes a node set $V$, an edge set $E$ and a adjacency matrix $A$. The adjacency matrix is an $N \times N$ square matrix, and the element $A_{ij}$ represents the connection relationship between node $v_i$ and node $v_j$. Traditional graph structures typically represent pair-wise relationships, where each edge connects two nodes. This design limits their ability to model many-to-many spatial interactions. To overcome this limitation, we adopt a hypergraph structure to represent spatial features, enabling the model to capture more complex and higher-order interactions. The structures of traditional graphs and hypergraphs are shown in Fig. 1.

A hypergraph is defined as $G=(V,E,H,W)$, which includes a node set $V$, a hyperedge set $E$, an incidence matrix $H$ and a hyperedge weight matrix $W$. In a hypergraph, each hyperedge is a subset of the nodes, allowing it to describe associations among multiple nodes rather than just pair-wise relationships. As shown in Fig. 1(b), hyperedge $e_1$ includes three nodes: $\{v_1, v_2, v_7\}$. Hypergraphs can therefore construct extensive node neighborhoods using a limited number of hyperedges. Additionally, since a node in a hypergraph can belong to multiple hyperedges, the dissemination of information within hypergraphs is more flexible and efficient compared to traditional graph structures. When each hyperedge connects only two nodes, hypergraph becomes equivalent to traditional graph structure. Therefore, hypergraph structure can be regarded as a generalized form of graph structure.

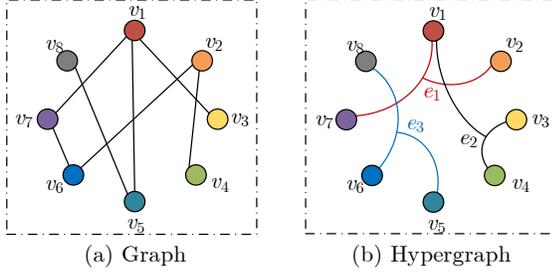

Fig. 1. The comparison between graph and hypergraph structure.

The connectivity relationship of hypergraph can be described using the incidence matrix $H$, defined as follows:

$$H_{ij} = \begin{cases} 1, v_i \in e_j \\ 0, v_i \notin e_j \end{cases}, H \in \mathbb{R}^{N \times I} \quad (3)$$

where $I$ is the number of hyperedges; the element $H_{ij}=1$ indicates that node $v_i$ belongs to hyperedge $e_j$, while $H_{ij}=0$ indicates that node $v_i$ does not belong to hyperedge $e_j$.

Additionally, a node degree diagonal matrix $D_e \in \mathbb{R}^{N \times N}$ and a hyperedge degree diagonal matrix $D_v \in \mathbb{R}^{I \times I}$ are defined to represent the degree of connectivity within the hypergraph as:

$$D_e(i,i) = \sum_j H_{i,j} \ ; \ D_v(i,i) = \sum_j W_j H_{i,j} \quad (4)$$

## III. SPATIO-TEMPORAL DYNAMIC HYPERGRAPH LEARNING

### A. Model Structure

The STDHL model is designed to establish a wind farm power forecasting framework using a hypergraph structure, as shown in Fig. 2.

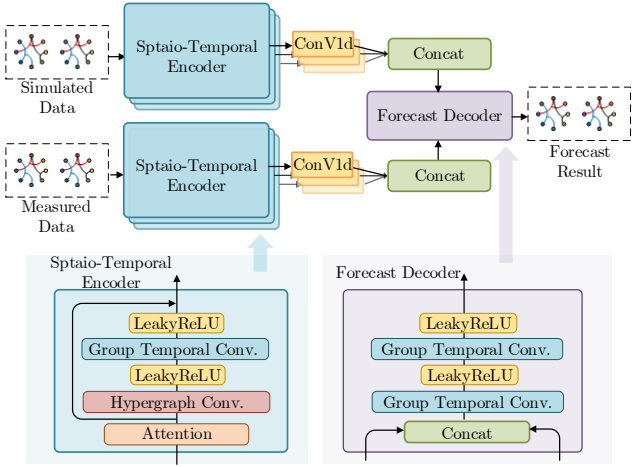

Fig. 2. The STDHL structure.

This hypergraph-based structure facilitates the fusion of measured and NWP data while flexibly capturing spatial features across multiple time scales. To achieve this, the model uses a pair of spatio-temporal encoders to map covariate data into a latent space. Each spatio-temporal encoder incorporates an attention layer to filter input data, extracting spatio-temporal features through stacked hypergraph convolutional layers and grouped temporal convolutional layers. The encoded data is subsequently concatenated after undergoing 1-dimensional convolutional channel adjustments. Finally, a forecasting decoder module integrates information from various modalities to produce the forecast results. Detailed descriptions of the model components are provided in the following subsections.

### B. Dynamic Hyperedge Convolution Layer

Hypergraph convolution mechanism can be inferred from spectral domain graph convolution theory as [30]:

$$g_\theta * \boldsymbol{x} = F^{-1}(F(g_\theta) \circ F(\boldsymbol{x})) \quad (5)$$

where $\boldsymbol{x}$ is hypergraph data; $g_\theta$ is the convolution kernel; $\circ$ is the element-wise Hadamard product; $F(\cdot)$ and $F^{-1}(\cdot)$ are the graph Fourier transform and inverse Fourier transform, respectively.

To avoid the computational burden of eigenvalue decomposition of the Laplacian matrix, Defferrard et al. [31] proposed transforming the graph convolution kernel into a Chebyshev polynomial function:

$$g_\theta * x = \sum_{k=0}^{K-1} \theta_k T_k(\hat{L})\boldsymbol{x} \quad (6)$$

where $T_k(\cdot)$ is the Chebyshev polynomial function; $T_0(a)=1$, $T_k(a) = 2aT_{k-1}(a) - T_{k-2}(a)$, and $\hat{L}=2L/\lambda_{\max} - I_n$; $L$ is the normalized hypergraph Laplacian matrix as [32]:

$$L = I_N - D_v^{-1/2} H W_e D_e^{-1/2} H^T D_v^{-1/2} = U \Lambda U^T \quad (7)$$

where $I_N \in \mathbb{R}^{N \times N}$ is the identity matrix; $\Lambda$ is the eigenvalues diagonal matrix of Laplacian matrix; $U$ is the eigenvector matrix of Laplacian matrix.

In traditional graph structures, applying a first-order Chebyshev truncation results in the loss of higher-order features. In contrast, for hypergraphs, first-order truncation of Chebyshev polynomials can be performed without concern, as hyperedges inherently capture higher-order feature information. Under these conditions, Eq. (6) can be simplified as follows:

$$g_\theta * \boldsymbol{x} \approx \theta_0 \boldsymbol{x} + \theta_1 D_v^{-1/2} H W_e D_e^{-1/2} H^T D_v^{-1/2} \boldsymbol{x} \quad (8)$$

The following parameter conversions are performed:

$$\begin{cases} \theta_1 = -\dfrac{1}{2}\theta \\ \theta_0 = \dfrac{1}{2}\theta D_v^{-1/2} H W_e D_e^{-1/2} H^T D_v^{-1/2} \end{cases} \quad (9)$$

Substituting Eq. (9) into Eq. (8) gives:

$$g_\theta * \boldsymbol{x} \approx \theta D_v^{-1/2} H W_e D_e^{-1/2} H^T D_v^{-1/2} \boldsymbol{x} \quad (10)$$

The above hypergraph can be equated to a two-stage convolution process as [30]:

$$\begin{cases} \boldsymbol{x}' = W_e D_e^{-1/2} H^T D_v^{-1/2} \boldsymbol{x} \\ \boldsymbol{x}'' = \theta D_v^{-1/2} H \boldsymbol{x}' \end{cases} \quad (11)$$

where $\boldsymbol{x}' \in \mathbb{R}^{I \times T}$ is the hyperedge feature and $\boldsymbol{x}'' \in \mathbb{R}^{N \times T}$ is the node feature. The information transfer process in hypergraph convolution can be understood as a two-step: first, information is passed from nodes to hyperedges, and then from hyperedges back to nodes, as shown in Fig. 3.

Calculating hypergraph convolution in Eq. (10) requires the incidence matrix and hyperedge weight matrix from prior information, which presents two challenges: (1) determining how to construct hyperedges and assign weights effectively to enhance forecast accuracy, and (2) accounting for the limitations of a static weight matrix in capturing the dynamic spatio-temporal correlations of wind power.

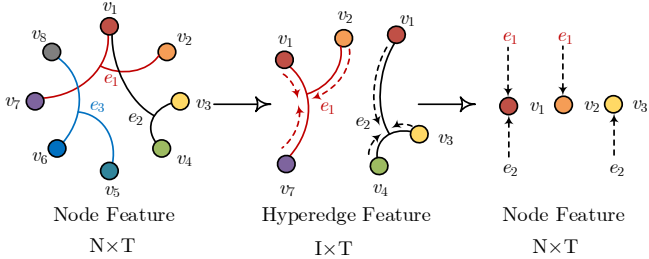

Fig. 3. The information transfer process of hypergraph convolution.

To address this, a dynamic hypergraph convolutional machine is proposed that uses two learnable matrices, $H_\theta^e(\boldsymbol{x}) \in \mathbb{R}^{N \times E}$ and $H_\theta^n(\boldsymbol{x}) \in \mathbb{R}^{E \times N}$, enabling a two-stage process of hypergraph convolution as:

$$\begin{cases} \boldsymbol{x}' = H_\theta^e(\boldsymbol{x}) \cdot \boldsymbol{x} \\ \boldsymbol{x}'' = H_\theta^n(\boldsymbol{x}) \cdot \boldsymbol{x}' \end{cases} \quad (12)$$

The learnable matrices are constructed as follows:

$$\begin{cases} d_{ij} = \| C_n(\boldsymbol{x})_i - C_e(\boldsymbol{x})_j \|_2 \\ H_\theta(i.j) = e^{-d_{ij}} / \sum_j e^{-d_{ij}} \end{cases} \quad (13)$$

where $C_n(\cdot)$ and $C_e(\cdot)$ is encode layer to transform the spatial dimensions of the hypergraph data; $d$ is the Euclidean distance under saptial dimensions; $H_\theta$ are computed using the Softmax function.

The hypergraph convolutional layer can be understood as the multiplication of graph data by a transfer matrix $H_\theta^n(\boldsymbol{x}) \cdot H_\theta^e(\boldsymbol{x})$. In the standard graph convolutional layer, this transfer matrix corresponds to the normalized Laplacian matrix. The transfer matrix serves as the link between graph convolution and hypergraph convolution, with the primary distinction being the method used to construct the transfer matrix, which in turn influences its capacity to capture higher-order correlations.

However, when performing the task of wind power forecasting, the hypergraph convolution has another drawback. It transmits only hyperedge features, which represent shared attributes among nodes. Consequently, Eq. (12) neglects the independent features of individual nodes during the convolution process. In ultra-short-term wind power forecasting, forecast results are not solely influenced by regional trends but are also strongly affected by the covariates specific to each node. To address this, a dynamic hypergraph convolutional layer is proposed, which incorporates self-loop feature transfer to account the nodes independent features:

$$\boldsymbol{x}'' = \alpha \cdot H_\theta^n(\boldsymbol{x}) \cdot H_\theta^e(\boldsymbol{x}) \cdot \boldsymbol{x} + (1-\alpha) \cdot \boldsymbol{x} \quad (14)$$

where $\alpha$ is a learnable weight coefficient.

### C. Group Temporal Convolutional Layer

Temporal convolutional networks (TCN) are effective tools for extracting temporal features [33]. Traditional TCN extract local features through weight sharing, but applying same temporal convolution kernel across all nodes in spatio-temporal forecasts limits the ability to capture variability among different farms. To address this, a grouped temporal convolution layer is proposed, which partitions the graph data into multiple independent groups for convolution. This method enhances the independence of temporal feature extraction across different nodes, enhancing the ability to capture unique patterns at each node. The spatial features are transferred exclusively through the hypergraph convolutional layer. The structure of group temporal convolutional as shown in Fig. 4.

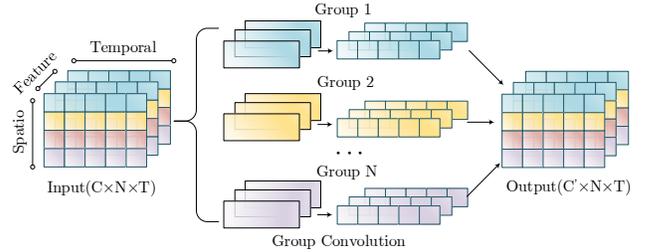

Fig. 4. Group temporal convolutional structure.

In a TCN, given an input $\boldsymbol{x} \in \mathbb{R}^{C \times N \times T}$ with $C$ channels. After temporal convolution, the output is transformed to $C'$ channels using a convolutional kernel $\boldsymbol{k} \in \mathbb{R}^{C \times C' \times \Delta T}$, where $\Delta T$ is the width of the temporal convolution kernel. In grouped TCN, data is divided based on the spatial dimension, represented by $\boldsymbol{x}(i,:) \in \mathbb{R}^{C \times T}$. Each node has its own independent temporal convolutional kernel $\boldsymbol{k}_i \in \mathbb{R}^{C \times C' \times \Delta T}$, allowing for customized temporal processing per node. The final output is obtained by concatenating the outputs of each group. The convolution formula for group TCN is as follows:

$$\boldsymbol{x}(c,i,t) = \sigma\left(\sum_{c'=1}^{C'} \sum_{w=W_1}^{W_{\Delta T}} \boldsymbol{k}_i(c,c',w)\boldsymbol{x}(i,t+dw) + b\right) \quad (15)$$

where $b$ is the bias, $\sigma$ is the activation function, and $d$ is the dilation rate.

In the wind forecast problem, causal convolution ($w \geq 0$) is used to encode measured data, ensuring that only current and past observations are incorporated into the model. Conversely, non-causal convolution is applied when encoding NWP data, as it does not require a restricted convolution parameter $w$, thereby enabling access to past, present, and future data points.

### D. Spatio-Temporal Encoder

The primary function of the spatio-temporal encoder is to integrate the hypergraph convolutional layer with the grouped temporal convolutional layer for effective spatio-temporal feature extraction from the covariates. To refine the feature selection, a location-based attention module is used to dynamically filter covarite features. The attention mechanism applies these weights through a Hadamard product with the original data, formulated as follows:

$$\boldsymbol{x}_{out} = \text{Softmax}(\sigma(\boldsymbol{x}_{in} * W + b)) * \boldsymbol{x}_{in} \quad (16)$$

After the attention-based feature filtering, the refined information is passed sequentially through the hypergraph convolutional layer and grouped temporal convolutional layers. To maintain feature quality across layers and prevent network degradation, the residual connection is added to the output of the spatio-temporal encoder.

*E. Forecast Encoder*

Mapping encoded covariate information to forecasts is a critical step in wind power forecasting. The forecast decoder is designed to integrate encoded data and generate forecast results. In this module, the encoded of measured data and NWP data are concatenated. Let $\boldsymbol{x}_m$ and $\boldsymbol{x}_n$ represent the encoded measured and NWP data, respectively. By concatenating them along the channel dimension, the concatenated data $\boldsymbol{x}_f = [\boldsymbol{x}_m, \boldsymbol{x}_n]$ is then processed through a grouped temporal convolutional layer, which adjusts the temporal dimensions to align with the target forecast dimensions and the channel dimensions to accommodate the required quantile forecasts. During training, the STDHL model is optimized by calculating the quantile loss between the forecasted and actual values as:

$$loss_{u,t} = \frac{1}{T}\sum_{t=t_0+1}^{t_0+T} \phi_t^u, \phi_t^u = \begin{cases} u(\boldsymbol{y}_t - \widehat{\boldsymbol{y}}_t^u), & \boldsymbol{y}_t \geq \widehat{\boldsymbol{y}}_t^u \\ (u-1)(\boldsymbol{y}_t - \widehat{\boldsymbol{y}}_t^u), & \boldsymbol{y}_t < \widehat{\boldsymbol{y}}_t^u \end{cases} \quad (17)$$

where $u$ is the quantile value; $\widehat{\boldsymbol{y}}_t^u$ is the $u$-quantile forecasts; $\phi_t^u$ is an indicative function.

## IV. CASE STUDY

*A. Data Description and Experiment Settings*

This dataset originates from the wind power forecasting track of the 2014 Global Energy Prediction Competition 2014. It includes measured power data and 24-hour-ahead NWP data for 10 wind farms in Australia [34]. The exact locations of these wind farms are not disclosed. The wind power data in the dataset is normalized as unit values, while the raw NWP data consists of both longitudinal and meridional wind components measured at two heights (10 m and 100 m). For experimental purposes, the raw wind data was converted to wind speed and further decomposed into sinusoidal and cosine components to represent wind direction.

The dataset spans from January 1, 2012, to December 31, 2013, with hourly temporal resolution. Of the total data, 70% was allocated for training, 10% for validation, and the remaining 20% for testing. The case study focuses on an ultra-short-term forecasting task for 10 wind farms next 4 hours. Unless otherwise specified, the look-back time horizon is set to 12 hours, and the extended NWP forecast horizon is 4 hours. The number of hyperedges is typically smaller than the number of nodes. In this study, an empirical value of $0.5N$ is adopted.

All experiments were conducted on a computer equipped with an NVIDIA TITAN V GPU and an Intel Core i9-7900X CPU. The models were implemented using PyTorch 1.8.0.

*B. Benchmark Models*

To systematically evaluate the proposed model, three categories of models were selected for comparative analysis:
*(1) Baseline Models*

Persistence Method (PSS) assumes that the power output will remain constant over the ultra-short-term, relying on the inertia of wind power. This model uses only measured data. Despite its simplicity, studies have shown that PSS provides valuable forecasts, particularly for immediate subsequent time steps.

Mechanism Model (MM) uses NWP wind speed values at 100 meters above ground level and applies the General Electric 1.5 MW wind turbine power curve for power conversion [35]. MM relies exclusively on NWP data.

Linear Model uses three fully connected layers to establish a baseline for feature extraction, offering reference results for evaluating more complex deep learning models.

Light gradient boosting machine (LightGBM) is an enhanced version of gradient-boosted decision trees, which has shown strong performance in various machine learning competitions [36]. It incorporates feature engineering within each quantile regressor to improve forecast accuracy.

*(2) State-of-the-art Deep Learning Models*

Deep autoregressive (DeepAR) uses an autoregressive recurrent neural network to model parametric Gaussian distribution, using the expected value and variance as probabilistic forecast results [37].

Improved density mixture density network (IDMDN) constructs a mixture density network through dense layers, using Beta distribution as its sub-distribution to prevent density leakage [38].

Transformer have gained significant attention in recent years, particularly in the development of large language models [39]. These architectures leverage positional encoding and self-attention mechanisms to effectively capture long-range dependencies in time-series data. The measured power data is fed into the Transformer encoder, while NWP data is fed into the Transformer decoder for feature extraction.

Patch time series transformer (PatchTST) is a specialized Transformer variant tailored for time-series forecasting. It segments the input time series into patches and applies self-attention mechanisms to simultaneously capture local patterns and long-term dependencies [40]. In this study, a time horizon of 3 hours is chosen for patch segmentation, and the encoder and decoder inputs are aligned with the Transformer.

*(3) STDHL Model Variants*

To further evaluate the proposed model, three STDHL variant models are introduced for comparison. In these variants, only the graph convolution layer is modified, while the remaining modules are kept unchanged:

Spatial-temporal static graph learning (STSGL) constructs a normalized Laplacian matrix using Pearson correlation coefficients derived from historical data. It uses a first-order approximation graph convolutional network to model spatial features.

Spatial-temporal dynamic graph learning (STDGL) uses a fully connected network to dynamically encode covariates, generating a normalized Laplacian matrix that reflects evolving spatial correlations. Similar to STSGL, it uses a first-order approximation graph convolutional network.

Spatial-temporal static hypergraph learning (STSHL) derives a fixed incidence matrix by clustering historical data using Gaussian mixture models, and it uses a unit matrix for the hyperedge weight matrix.

*C. Evaluation Metrics*

Six evaluation metrics are used in this study, encompassing both deterministic and probabilistic metrics. For the probabilistic forecasting models, the quantile model uses the median, while the probability density function model uses the expected value as their deterministic forecasting results.

MAE and RMSE are used as evaluation metrics for deterministic forecasts:

$$\text{MAE} = \mathbb{E}_{s,t} \mid \boldsymbol{y}_{s,t} - \widehat{\boldsymbol{y}}_{s,t} \mid \quad (18)$$

$$\text{RMSE} = \sqrt{\mathbb{E}_{s,t}(\boldsymbol{y}_{s,t} - \widehat{\boldsymbol{y}}_{s,t})^2} \quad (19)$$

where $\boldsymbol{y}_{s,t}$ is the measured power value of the $s$-th wind farm at moment $t$, $\widehat{\boldsymbol{y}}_{s,t}$ is the forecast results of the $s$-th wind farm at moment $t$.

The accuracy rate (AR) and the pass rate (PP) are used as evaluation metrics for wind power forecast results. In China, only the fourth hour forecast is evaluated by AR and PP, where the baseline for both AR and PP is 87%, and wind farms face fines when they fall below the baseline.

$$\text{AR} = [1 - \text{RMSE}_{t_0+4}] \times 100\% \quad (20)$$

$$\text{PP} = \mathbb{E}_s(B_s) \times 100\%; B_s = \begin{cases} 1 & \mid y_{t_0+4} - \widehat{y}_{t_0+4} \mid < 0.25 \\ 0 & \mid y_{t_0+4} - \widehat{y}_{t_0+4} \mid \geq 0.25 \end{cases} \quad (21)$$

AR corresponds to the fourth hour RMSE, while PP evaluates the percentage of the fourth hour MAE that is less than a set threshold.

The pinball score (PS) and continuous ranked probability score (CRPS) are used as evaluation metrics for probabilistic forecasts. The PS is calculated as in Eq. (15).

$$\text{CRPS} = \mathbb{E}_{s,t}[\mathbb{E}_m \mid \boldsymbol{s}_{s,t}^m - \boldsymbol{y}_{s,t} \mid - \mathbb{E}_{m,n} \mid \boldsymbol{s}_{s,t}^m - \boldsymbol{s}_{s,t}^n \mid] \quad (22)$$

where $\boldsymbol{s}_{s,t}^m$ and $\boldsymbol{s}_{s,t}^n$ are independent samples sampled from the forecast probability distribution $p(\widehat{\boldsymbol{y}}_{s,t})$.

## V. RESULTS AND DISCUSSION

*A. Forecast Results*

This section evaluates the performance of the benchmark models in comparison to the STDHL model for ultra-short-term wind power forecasting. Table I provides a summary of the results across six evaluation metrics. To assess the forecasting accuracy of different models over varying time horizons, Fig. 5 presents the MAE for forecast from 1-hour to 4-hour ahead. By analyzing the results presented in Table 1 and Fig. 5, the following conclusions can be drawn.

(1) The MAE and RMSE of the PSS outperform those of the MM at 1-hour to 3-hour ahead forecasts. The MAE of MM is much higher than any other model in the 1-hour ahead forecast. This is largely influenced by the timeliness of the NWP. Notably, the MAE of PSS ranks third only to STDHL and IDMDN for 1-hour ahead forecasts. However, the forecast error of PSS increases progressively as the forecast horizon extends, whereas MM maintains a more stable trend. For 4-hour ahead forecasts, the MAE of MM surpasses that of PSS. Nevertheless, both AR and PP values fall short of the baseline of 87%, suggesting that relying solely on measured data or NWP data is insufficient to meet the assessment requirements.

(2) Apart from PSS and MM, the remaining seven models integrate both measured data and NWP data, resulting in superior performance across all metrics compared to PSS and MM. The proposed STDHL model achieves the best performance across all evaluation metrics. PatchTST records the second-best AR score, while IDMDN records second-best in the remaining metrics. Compared to these second-best scores STDHL demonstrates the following improvements: MAE by 9.96%, RMSE by 8.63%, AR by 1.25%, PP by 2.45%, CRPS by 5.41%, and PS by 5.58%. Additionally, DeepAR, PatchTST, IDMDN, and STDHL achieve AR and PP values of 87%, thereby meeting the assessment requirements.

(3) Although Transformer-based architectures are often regarded as state-of-the-art, their performance in this study is less remarkable. For instance, in 1-hour ahead forecasts, the MAE of Transformer and PatchTST is higher than that of PSS. Self-attention mechanisms in Transformers assign weights to specific time horizon, which are effective for time series with strong periodic patterns. However, the periodicity of wind power is less pronounced, which diminishes the effectiveness of this approach. Furthermore, the basic Transformer is limited capacity to model fine-grained spatial features may lead to its weaker performance in spatio-temporal forecasting tasks.

TABLE I
COMPARISON OF THE FORECAST RESULTS

| Model | MAE | RMSE | AR | PP | CRPS | PS |
|---|---|---|---|---|---|---|
| PSS | 0.1216 | 0.1380 | 83.61 | 76.72 | - | - |
| Mechanism | 0.1574 | 0.1748 | 84.22 | 77.88 | - | - |
| Linear | 0.1157 | 0.1299 | 85.12 | 81.53 | - | - |
| LightGBM | 0.1025 | 0.1165 | 87.24 | 84.42 | 0.0743 | 0.0389 |
| DeepAR | 0.1018 | 0.1145 | 88.20 | 88.29 | 0.0743 | 0.0389 |
| Transformer | 0.1061 | 0.1202 | 87.75 | 86.70 | 0.0784 | 0.0410 |
| PatchTST | 0.1010 | 0.1150 | **88.46** | 88.25 | 0.0736 | 0.0384 |
| IDMDN | 0.0984 | 0.1113 | 87.99 | 88.42 | 0.0684 | 0.0358 |
| STDHL | **0.0886** | **0.1017** | **89.57** | **90.59** | **0.0647** | **0.0338** |

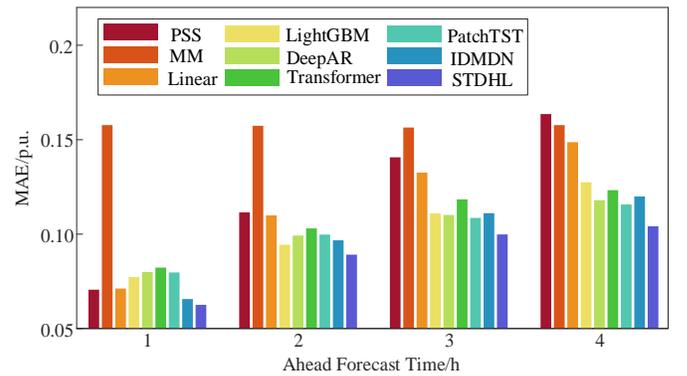

Fig.5 The MAE of 1-hour to 4-hour ahead forecast results.

To further illustrate the forecasting performance, Fig. 6 presents the forecasting results for two consecutive days, including forecasts from 1-hour to 4-hour ahead. The following conclusions can be drawn.

(1) As the forecast ahead time increases, the confidence interval of the STDHL model gradually widens, indicating an increase in uncertainty. The PSS forecasts are close to the measured values for the 1-hour ahead forecast. However, when wind power experiences significant fluctuations, the forecast error of the PSS model increases sharply. By the 4-hour ahead forecast, the PSS model is unable to effectively capture the measured wind power fluctuation trends. In comparison, the MM model exhibits more stability, with forecast results that do not change with the lead time. This is because the NWP data used in the MM model is obtained before the day and is not updated on a rolling basis.

(2) The complementarity of various covariate data is clearly demonstrated. The PSS can capture the volatility of wind power in the present, NWP data effectively tracks the fluctuation trends of wind power but is prone to amplitude and timing biases. During the 6:00 to 18:00 period shown in Fig. 6, the MM model successfully forecasts the wind power ramp-up but shows a noticeable lag in timing. In contrast, the proposed STDHL model, which integrates both measured and NWP data, provides more accurate estimations of wind power fluctuations. This integration significantly enhances forecast accuracy and effectively addresses the limitations of using a single covariate.

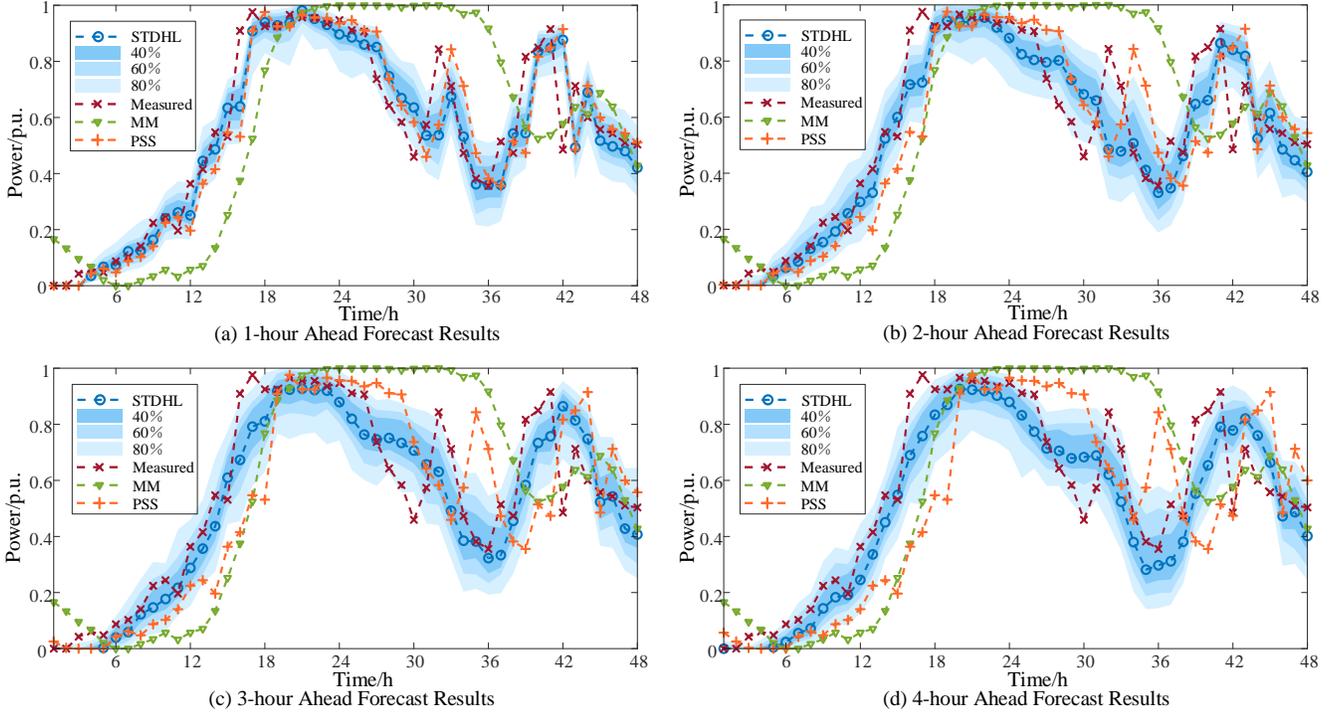

Fig.6 Forecast results of ultra-short-term wind power.

*B. Graph Analysis*

In this section, four different graph models are compared to illustrate the superiority of the proposed dynamic hypergraph convolution mechanism. The forecast results as shown in Table II. In addition, in order to compare the spatial transfer feature of the different graph models, the heatmap of transfer matrices as shown in Fig. 7.

TABLE II
COMPARISON OF THE GRAPH FORECAST MODEL

| Model | MAE | RMSE | AR | PP | CRPS | PS |
|---|---|---|---|---|---|---|
| STSGL | 0.0928 | 0.1061 | 89.12 | 89.43 | 0.0676 | 0.0353 |
| STDGL | 0.0916 | 0.1035 | 89.24 | 89.94 | 0.0662 | 0.0346 |
| STSHL | 0.1006 | 0.1135 | 88.65 | 88.44 | 0.0731 | 0.0382 |
| STDHL | 0.0886 | 0.1017 | 89.57 | 90.59 | 0.0647 | 0.0338 |

As can be shown in Table 2, the STDHL model performs the best forecast accuracy, followed by STDGL and STSGL. however, the STSHL method has a lower accuracy in ultra-short-term wind power forecasting than the traditional static and dynamic graph model.

The effectiveness of the graph model in capturing spatial features is more clearly demonstrated through the transfer matrix analysis. As shown in Fig. 7(a), the majority of transfer matrix elements in the STSGL model exceed 0.5. This result is primarily influenced by the Pearson correlation coefficient, which restricts the flexibility of representing spatial correlations among wind farms.

In contrast, Fig. 7(b) shows that the STDGL model not only captures strong correlations within node self-information but also constructs an asymmetric transfer matrix that dynamically represents spatial correlations. This dynamic adaptability provides a notable advantage over the STSGL model.

Fig. 7(c) shows that the diagonal values of the STSHL model are relatively lower. This outcome arises because, when the number of hyperedges is smaller than the number of nodes, the hypergraph prioritizes higher-order interactions among multiple nodes, thereby reducing emphasis on individual node self-information.

Finally, Fig. 7(d) shows the ability to learn asymmetric transfer matrices from covariates while retaining node self-information. This balance between dynamic spatial correlation modeling and self-information preservation enables the STDHL model to outperform traditional hypergraph structures in ultra-short-term wind power forecasting.

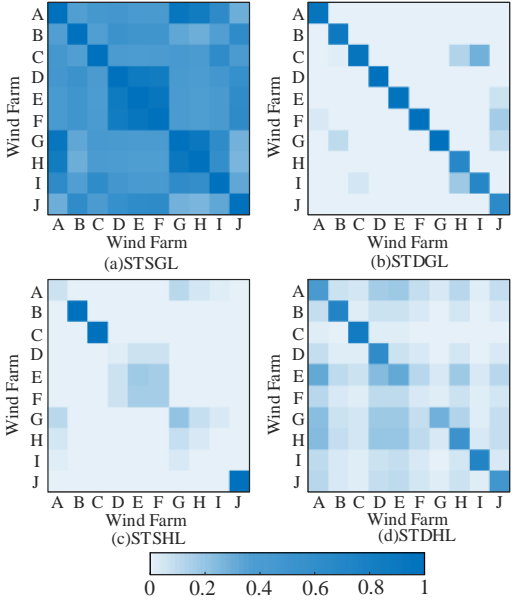

Fig.7 Transfer matrix heatmap of different graph model.

## C. Spatio-Temporal Covariate Ablation experiment

In this section, ablation experiments are conducted on the spatio-temporal covariate data to analyze their impact on forecasting accuracy. For the temporal dimension, the experiments involve varying the look-back time horizon of measured data and the extended time horizon of NWP data to perform a comparative analysis of the forecasting results. In the spatial dimension, the number of wind farms included in the input data is incrementally increased, with farms A and B designated as target farms, to evaluate the effect of spatial data on forecast accuracy.

Fig. 8 and 9 show the forecast results for varying look-back time horizons of measured data and extended time horizons of NWP data, respectively. Fig. 10 presents the forecast results corresponding to different numbers of wind farms included in the covariate data. Fig. 8 shows that extending the look-back time horizon improves ultra-short-term forecast accuracy. However, once the time horizon exceeds 12 hours, the forecast accuracy tends to stabilize. The extended look-back time horizon effectively captures real-time power fluctuations and, when combined with NWP data, aids in trend forecast and the correction of deviations. Fig. 9 shows that extending the time horizon of NWP data enhances forecast accuracy, with the highest accuracy observed at a 4-hour extension. This finding aligns with the observations in Fig. 6. Due to inherent deviations in NWP data, restricting the covariate selection to only the target forecast time horizon may result in the loss of valuable NWP trend information.

Fig. 10 shows the impact of spatial data on wind power forecasting accuracy. While incorporating additional spatial data generally enhances accuracy, the degree of improvement depends on the correlation between the spatial data and the target wind farm. For example, wind farm B showed a gradual decline in forecast accuracy when data from 10 farms were integrated. In contrast, wind farm A experienced no notable improvement when data from the first six farms were included but exhibited a significant enhancement when data from wind farms G and H were added. This improvement is attributed to the stronger correlation between wind farm A and wind farms G and H, as verified by the heatmap in Fig. 7(a). These results underscore the importance of integrating spatial data from highly correlated wind farms to significantly enhance the accuracy of ultra-short-term wind power forecasting.

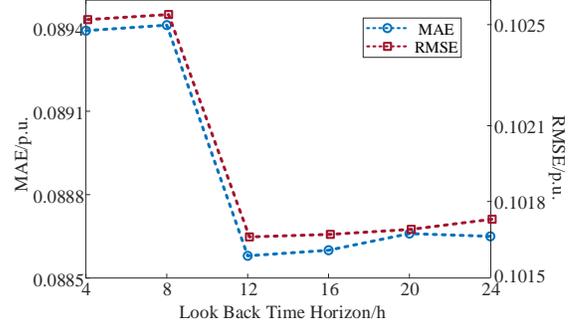

Fig.8 The MAE of measured data ablation experiment.

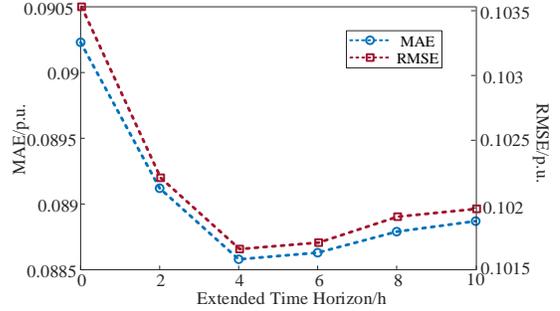

Fig.9 The MAE of NWP data ablation experiment.

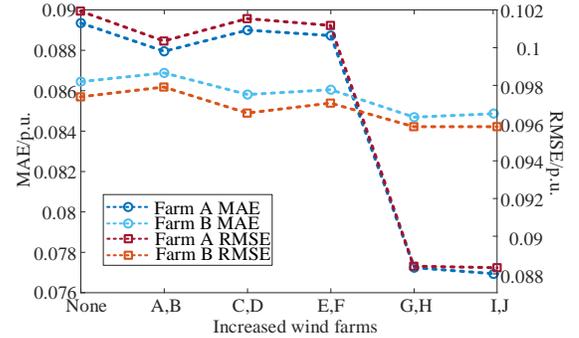

Fig.10 The MAE of spatial data ablation experiment.

## VI. CONCLUSION

In this study, an STDHL model is proposed for ultra-short-term wind power forecasting. The model introduces a hypergraph learning framework to describe the spatial relationships among wind farms. Notably, the hypergraph structure can capture complex higher-order spatial correlations between graph nodes, which is an ability lacking in traditional graph learning. Building on this framework, a novel hypergraph convolutional layer and a grouped temporal convolutional layer is designed for wind power forecasting. The STDHL model dynamically extracts spatial features across multiple time scales while preserving the independent temporal modeling of each wind farm.

The results of the case study demonstrate the superiority of the STDHL model in ultra-short-term wind power forecasting. Through detailed analysis, it is evident that self-loop mechanism is indispensable in the hypergraph forecasting model. Furthermore, ablation experiments on spatio-temporal covariates highlight the significant advantages of integrating measured and NWP data from multiple wind farms. And the findings reveal that extending the NWP time horizon substantially mitigates the impact of time bias.

Future research will focus on addressing challenges related to partial node data corruption or missing data in large-scale spatio-temporal datasets. In particular, efforts will aim to develop hypergraph forecasting models with plug-and-play functionality for spatio-temporal covariates, facilitating advancements in the practical application of spatio-temporal forecasting.

## REFERENCES

[1] D. Neves, M. C. Brito, and C. A. Silva, "Impact of solar and wind forecast uncertainties on demand response of isolated microgrids," *Renew. Energy*, vol. 87, pp. 1003–1015, 2016.
[2] X. Dong, Z. Mao, Y. Sun, and X. Xu, "Short-Term Wind Power Scenario Generation Based on Conditional Latent Diffusion Models," *IEEE Trans. Sustain. Energy*, vol. 15, no. 2, pp. 1074–1085, Apr. 2024.
[3] F. Díaz-González, M. Hau, A. Sumper, and O. Gomis-Bellmunt, "Participation of wind power plants in system frequency control: Review of grid code requirements and control methods," *Renew. Sustain. Energy Rev.*, vol. 34, pp. 551–564, 2014.
[4] S. S. Soman, H. Zareipour, O. Malik, and P. Mandal, "A review of wind power and wind speed forecasting methods with different time horizons," *North Am. Power Symp. 2010, NAPS 2010*, pp. 1–8, 2010.
[5] D. W. Gao, E. Muljadi, T. Tian, *et al.* "Comparison of standards and technical requirements of grid-connected wind power plants in China and the United States". *National Renewable Energy Lab.*, (No. NREL/TP-5D00-64225), 2016.
[6] W. Wang *et al.*, "Accurate initial field estimation for weather forecasting with a variational constrained neural network," *npj Clim. Atmos. Sci.*, vol. 7, no. 1, pp. 1–17, 2024.
[7] A. Gettelman et al., "The future of Earth system prediction: Advances in model-data fusion," *Sci. Adv.*, vol. 8, no. 14, pp. 1–12, 2022.
[8] M. Yu, J. Han, H. Wu, J. Yan, and R. Zeng, "Short-Term Wind Power Prediction Based on Wind2vec-BERT Model," *IEEE Trans. Sustain. Energy*, vol. 81, no. 1, pp. 1–12, 2024.
[9] X. Dong, Y. Sun, Y. Li, *et al*, "Spatio-temporal Convolutional Network Based Power Forecasting of Multiple Wind Farms," *J. Mod. Power Syst. Clean Energy*, vol. 10, no. 2, pp. 388–398, 2022.
[10] P. Bauer, A. Thorpe, and G. Brunet, "The quiet revolution of numerical weather prediction," *Nature*, vol. 525, no. 7567, pp. 47–55, 2015.
[11] Y. Li, H. Wang, J. Yan, C. Ge, S. Han, and Y. Liu, "Ultra-Short-Term Wind Power Forecasting Based on the Strategy of 'Dynamic Matching and Online Modeling,'" *IEEE Trans. Sustain. Energy*, vol. PP, pp. 1–16, 2024.
[12] M. Yang, Y. Huang, C. Xu, et al, "Review of several key processes in wind power forecasting: Mathematical formulations, scientific problems, and logical relations," *Appl. Energy*, vol. 377, p. 124631, 2025.
[13] H. Zhang, J. Yan, Y. Liu, Y. Gao, S. Han, and L. Li, "Multi-source and temporal attention network for probabilistic wind power prediction," *IEEE Trans. Sustain. Energy*, vol. 12, no. 4, pp. 2205–2218, 2021.
[14] Y. Wang, Q. Hu, D. Srinivasan, and Z. Wang, "Wind Power Curve Modeling and Wind Power Forecasting with Inconsistent Data," *IEEE Trans. Sustain. Energy*, vol. 10, no. 1, pp. 16–25, 2019.
[15] X. Liu, Z. Lin, and Z. Feng, "Short-term offshore wind speed forecast by seasonal ARIMA - A comparison against GRU and LSTM," *Energy*, vol. 227, p. 120492, 2021.
[16] M. Landry, T. P. Erlinger, D. Patschke, and C. Varrichio, "Probabilistic gradient boosting machines for GEFCom2014 wind forecasting," *Int. J. Forecast.*, vol. 32, no. 3, pp. 1061–1066, 2016.
[17] H. Wang, Z. Lei, X. Zhang, B. Zhou, and J. Peng, "A review of deep learning for renewable energy forecasting," *Energy Convers. Manag.*, vol. 198, no. April, p. 111799, 2019.
[18] Y. Yu, M. Yang, X. Han, Y. Zhang, and P. Ye, "A Regional Wind Power Probabilistic Forecast Method Based on Deep Quantile Regression," *IEEE Trans. Ind. Appl.*, vol. 57, no. 5, pp. 4420–4427, 2021.
[19] X. Liu, L. Yang, and Z. Zhang, "Short-term multi-step ahead wind power predictions based on a novel deep convolutional recurrent network method," *IEEE Trans. Sustain. Energy*, vol. 12, no. 3, pp. 1820–1833, 2021.
[20] F. Corradini, M. Gori, C. Lucheroni, M. Piangerelli, and M. Zannotti, "A Systematic Literature Review of Spatio-Temporal Graph Neural Network Models for Time Series Forecasting and Classification," 2024, [Online]. Available: http://arxiv.org/abs/2410.22377
[21] M. Khodayar and J. Wang, "Spatio-Temporal Graph Deep Neural Network for Short-Term Wind Speed Forecasting," *IEEE Trans. Sustain. Energy*, vol. 10, no. 2, pp. 670–681, Apr. 2019.
[22] X. Peng, Y. LI, and F. Tsung, "Wind Power Ramp Events Prediction Considering Wind Propagation," *Renew. Energy*, vol. 236, no. October 2023, p. 121280, 2024.
[23] Z. Li *et al.*, "A Spatiotemporal Directed Graph Convolution Network for Ultra-Short-Term Wind Power Prediction," *IEEE Trans. Sustain. Energy*, vol. 14, no. 1, pp. 39–54, Jan. 2023.
[24] Z. Li *et al.*, "Heterogeneous Spatiotemporal Graph Convolution Network for Multi-Modal Wind-PV Power Collaborative Prediction," *IEEE Trans. Power Syst.*, vol. 39, no. 4, pp. 5591–5608, 2024.
[25] J. Tang, Z. Liu, and J. Hu, "Spatial-Temporal Wind Power Probabilistic Forecasting Based on Time-Aware Graph Convolutional Network," *IEEE Trans. Sustain. Energy*, vol. 15, no. 3, pp. 1946–1956, 2024.
[26] F. Wang et al., "Dynamic spatio-temporal correlation and hierarchical directed graph structure based ultra-short-term wind farm cluster power forecasting method," *Appl. Energy*, vol. 323, no. March 2021, p. 119579, Oct. 2022.
[27] J. J. Q. Yu and J. Gu, "Real-Time Traffic Speed Estimation with Graph Convolutional Generative Autoencoder," *IEEE Trans. Intell. Transp. Syst.*, vol. 20, no. 10, pp. 3940–3951, 2019.
[28] Y. Zhu, Y. Zhou, W. Wei, and L. Zhang, "Real-Time Cascading Failure Risk Evaluation With High Penetration of Renewable Energy Based on a Graph Convolutional Network," *IEEE Trans. Power Syst.*, vol. 38, no. 5, pp. 4122–4133, 2023.
[29] S. Liu, L. Chen, H. Dong, Z. Wang, D. Wu, and Z. Huang, "Higher-order Weighted Graph Convolutional Networks," pp. 1–15, Nov. 2019, [Online]. Available: http://arxiv.org/abs/1911.04129
[30] J. Bruna, W. Zaremba, A. Szlam et al., "Spectral networks and locally connected networks on graphs," in *Proceedings of 2th International Conference on Learning Representations (ICLR)*, Banff, Canada, Dec. 2013, pp. 1-14.
[31] M. Defferrard, X. Bresson, and P. Vandergheynst, "Convolutional neural networks on graphs with fast localized spectral filtering," in *Proc. Int. Conf. Neural Inf. Process. Syst.*, 2016, Art. no. 29.
[32] Y. Gao, Y. Feng, S. Ji, and R. Ji, "HGNN+: General Hypergraph Neural Networks," *IEEE Trans. Pattern Anal. Mach. Intell.*, vol. 45, no. 3, pp. 3181–3199, 2023.
[33] J. Liang and W. Tang, "Ultra-Short-Term Spatiotemporal Forecasting of Renewable Resources: An Attention Temporal Convolutional Network-Based Approach," *IEEE Trans. Smart Grid*, vol. 13, no. 5, pp. 3798–3812, Sep. 2022.
[34] T. Hong, P. Pinson, S. Fan, H. Zareipour, A. Troccoli, and R. J. Hyndman, "Probabilistic energy forecasting: Global energy forecasting competition 2014 and beyond," *Int. J. Forecasting*, vol. 32, no. 3, pp. 896–913, 2016.
[35] Wind turbine models. [Online] https://en.wind-turbine-models.com/turbines/565-ge-general-electric-ge-1.5s
[36] G. Ke, Q. Meng, T. Finley, *et al.*, "LightGBM: A highly efficient gradient boosting decision tree," *Adv. Neural Inf. Process. Syst.*, pp. 3147–3155, 2017.
[37] D. Salinas, V. Flunkert, J. Gasthaus, and T. Januschowski, "DeepAR: Probabilistic forecasting with autoregressive recurrent networks," *Int. J. Forecast.*, vol. 36, no. 3, pp. 1181–1191, 2020.
[38] H. Zhang, Y. Liu, J. Yan, et al, "Improved deep mixture density network for regional wind power probabilistic forecasting," *IEEE Trans. Power Syst.*, vol. 35, no. 4, pp. 2549–2560, Jul. 2020.
[39] Ashish Vaswani, Noam Shazeer, Niki Parmar, Jakob Uszkoreit, Llion Jones, Aidan N Gomez, Lukasz Kaiser, and Illia Polosukhin. Attention is all you need. In *Advances in Neural Information Processing Systems*, volume 30, 2017.
[40] Y. Nie, N. H. Nguyen, P. Sinthong, and J. Kalagnanam, "A Time Series is Worth 64 Words: Long-term Forecasting with Transformers," Nov. 2022, [Online]. Available: http://arxiv.org/abs/2211.14730